\def\year{2022}\relax
\documentclass[letterpaper]{article} 
\usepackage{aaai22}  
\usepackage{times}  
\usepackage{helvet}  
\usepackage{courier}  
\usepackage[hyphens]{url}  
\usepackage{graphicx} 
\usepackage{natbib}  
\usepackage{caption} 
\DeclareCaptionStyle{ruled}{labelfont=normalfont,labelsep=colon,strut=off} 
\usepackage{algorithm}
\usepackage{algpseudocode}

\usepackage{newfloat}
\usepackage{listings}
\floatstyle{ruled}
\newfloat{listing}{tb}{lst}{}
\floatname{listing}{Listing}
\usepackage{amsmath}
\usepackage{array}
\usepackage{makecell}
\usepackage{multirow}
\usepackage{booktabs} 

\title{Improving Neural Cross-Lingual Abstractive Summarization via Employing Optimal Transport Distance for Knowledge Distillation}
\author{
    Thong Nguyen$^1$, Luu Anh Tuan$^2$\thanks{Corresponding author}
}
\affiliations{
    $^1$ VinAI Research, Vietnam \\
    $^2$ Nanyang Technological University, Singapore
}

\usepackage{bibentry}

\begin{document}

\maketitle

\begin{abstract}
Current state-of-the-art cross-lingual summarization models employ multi-task learning paradigm,  which works on a shared vocabulary module and relies on the self-attention mechanism to attend among tokens in two languages. However, correlation learned by self-attention is often loose and implicit, inefficient in capturing crucial cross-lingual representations between languages. The matter worsens when performing on languages with separate morphological or structural features, making the cross-lingual alignment more challenging, resulting in the performance drop. To overcome this problem, we propose a novel Knowledge-Distillation-based framework for Cross-Lingual Summarization, seeking to explicitly construct cross-lingual correlation by distilling the knowledge of the monolingual summarization teacher into the cross-lingual summarization student. Since the representations of the teacher and the student lie on two different vector spaces, we further propose a Knowledge Distillation loss using Sinkhorn Divergence, an Optimal-Transport distance, to estimate the discrepancy between those teacher and student representations. Due to the intuitively geometric nature of Sinkhorn Divergence, the student model can productively learn to align its produced cross-lingual hidden states with monolingual hidden states, hence leading to a strong correlation between distant languages. Experiments on cross-lingual summarization datasets in pairs of distant languages demonstrate that our method outperforms state-of-the-art models under both high and low-resourced settings.
\end{abstract}

\section{Introduction}

Cross-Lingual Summarization (CLS) is the task of condensing a document of one language into its shorter form in the target language. Most of contemporary works can be classified into two categories, i.e. low-resourced and high-resourced CLS approaches. In high-resourced scenarios, models are provided with an enormous number of document /summary pairs on which they can be trained \cite{zhu2019ncls, cao2020jointly, zhu2020attend}. On the other hand, in low-resourced settings, those document/summary pairs are scarce, which restrains the amount of information that a model can learn. While high-resourced settings are preferred, in reality it is difficult to attain a sufficient amount of data, especially for less prevalent languages. 
\begin{figure}[t]
\centering
\includegraphics[width=\linewidth]{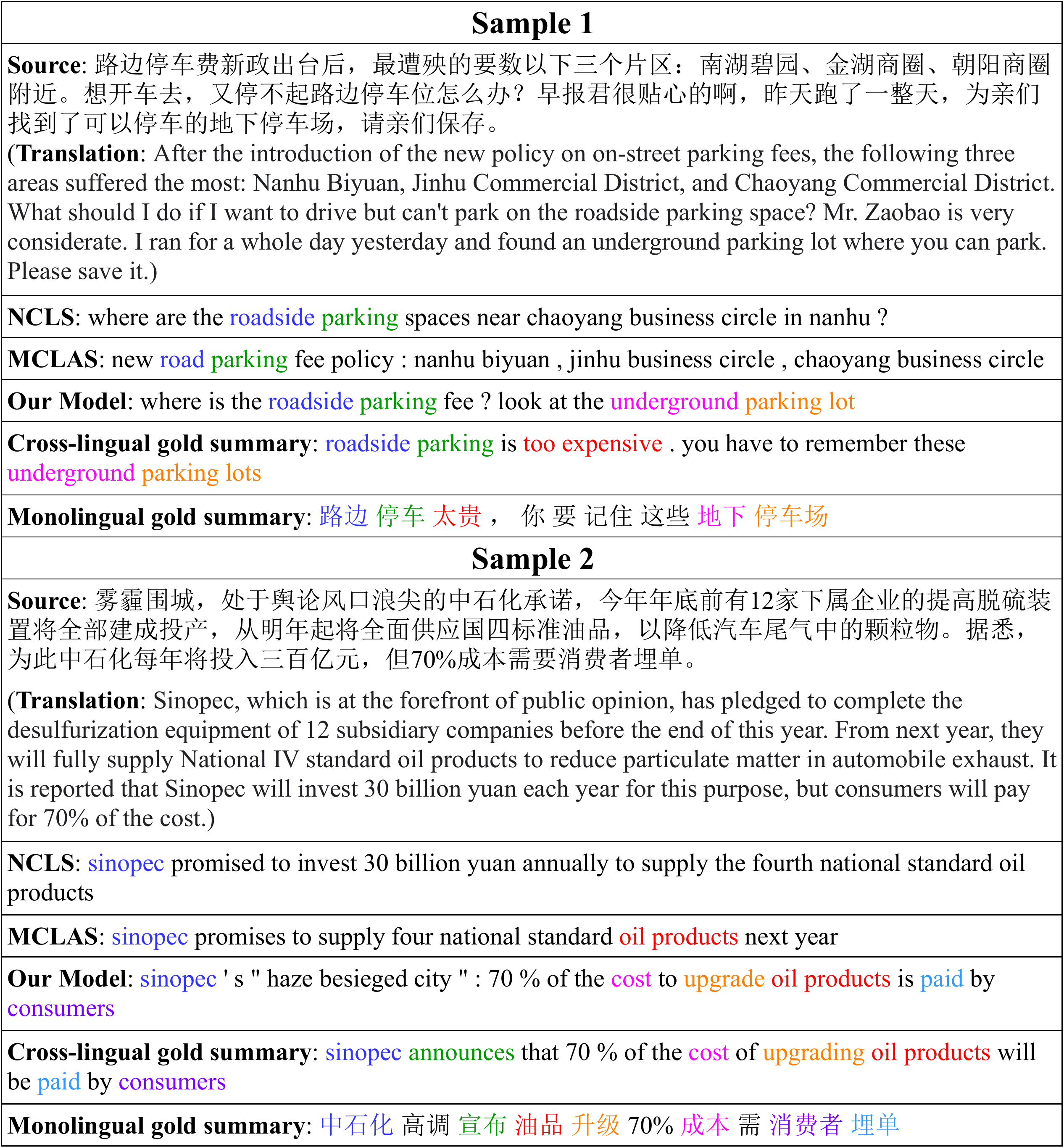}
\caption{Examples of Chinese-English cross-lingual summarization. Here we present the English output generated by the NCLS model of \cite{zhu2019ncls}}
\label{fig:example}
\end{figure}

Most previous works resolving the issue of little training data concentrate on multi-task learning framework by utilizing the relationship of Cross-Lingual Summarization (CLS) with Monolingual Summarization (MLS) or Neural Machine Translation (NMT). Their approach can be further divided into two groups. The first group equips their module with two independent decoders, one of them targets the auxiliary task (MLS or NMT). Nevertheless, since two decoders do not share their parameters, this approach undermines the model’s ability to align between two tasks \cite{bai2021cross}, making the ancillary and the main task less relied upon each other. Hence, the trained model might produce output that does not match up the topic, or miss important spans of text. In Figure \ref{fig:example}, we list two samples of documents with their gold and generated summaries by the model NCLS of \cite{zhu2019ncls}. As it can be seen, their cross-lingual outputs do not include key spans from the summary, e.g., ``\emph{underground}'', ``\emph{parking lot}'' in sample 1, and  ``\emph{consumers}'' in sample 2. In both samples, the content of the cross-lingual summary diverges significantly from the one of the monolingual gold summary.

The second group decides to employ a single decoder dealing with both CLS and MLS tasks. To this end, the method concatenates the monolingual to cross-lingual summary and designate the model to sequentially generate the monolingual summary, and then the cross-lingual one. Unfortunately, notwithstanding lessening the computational overhead during training by using solely one decoder, this method is not efficacious in capturing the connection between two languages in the output, consequently producing representations that do not take into account language relationships \cite{luoveco}. In that case, the correlation of cross-lingual representations will be tremendously impacted by the structural and morphological similarity of those languages \cite{bjerva2019language}.  As a result, in case of summarizing the document from one language to another that possesses distinct morphology and structure properties, such as from Chinese to English, the decoder might be prone to underperformance, due to the dearth of language correlation between two sets of hidden representations in the bilingual vector space \cite{luoveco}.

To solve the aforementioned problem, we propose a novel Knowledge-Distillation framework for Cross-Lingual Summarization task. Particularly, our framework consists of a teacher model targetting Monolingual Summarization, and a student for Cross-Lingual Summarization. We initiate our procedure by finetuning the teacher model on monolingual document/summary pairs. Subsequently, we continue to distill summarization knowledge of the trained teacher into the student model. Because the hidden vectors of the teacher and student lie upon two disparate monolingual and cross-lingual spaces, respectively, we propose a Sinkhorn-Divergence-based Knowledge Distillation loss, for the distillation process. Whereas multiple distances such as Cosine Distance or Euclidean Distance demand two sets share the sample size and are sensitive to outliers \cite{zimek2012survey}, Sinkhorn divergence does not enforce any requirement that relates to the number of samples and is also robust to noise \cite{sejourne2019sinkhorn}. Furthermore, compared with other types of divergences such as KL divergence, the computation of Sinkhorn divergence does not require two distributions to lie on the same probability space. This is important because two languages might possess distinct features that cannot be projected one-to-one, such as the vocabulary set. Consequently, employing divergences different from Sinkhorn would need additional constraint to the distillation loss. Lastly, Sinkhorn divergence is able to capture geometric nature \cite{feydy2019interpolating} which has been shown to benefit myriad cross-lingual and multilingual representation learning settings \cite{huang2021improving}. We will empirically prove the superiority of Sinkhorn divergence in the Experiment section. 

Since the proposed module perpetuates the one-decoder employment, our framework is able to explicitly correlate representations from two languages, thus resolving the issue of two distant languages without demanding any additional computation overhead. To evaluate the efficacy of our framework, we proceed to conduct the experiments on myriad datasets containing document /summary pairs of couples of distant languages, for example, English-to-Chinese, English-to-Arabic, Japanese-to-English, etc. The empirical results demonstrate that our model outperforms previous state-of-the-art Cross-Lingual Summarization approaches. In sum, our contributions are three-fold:
\begin{itemize}
    \item We propose a Knowledge Distillation framework for Cross-Lingual Summarization task, which seeks to enhance the summarization performance on distant languages by aligning the cross-lingual with monolingual summarization, through distilling the knowledge of monolingual teacher into cross-lingual student model.
    \item We propose a novel Knowledge Distillation loss using Optimal-Transport distance, i.e. Sinkhorn Diveregence, with a view to coping with the spatial discrepancy formed by the hidden representations produced by teacher and student model.
    \item We conducted extensive experiments in both high and low-resourced settings on multiple Cross-Lingual Summarization datasets that belong to pairs of morphologically and structurally distant languages, and found that our method significantly outperforms other baselines in both automatic metrics and by human evaluation.
\end{itemize}
\section{Related Work}

\subsection{Neural Cross-Lingual Summarization}

Due to the advent of Transformer architecture with its self-attention mechanism, Text Generation has received ample attention from researchers \cite{tuan2020capturing, lyu2021improving, zhang2021data}, especially Document Summarization \cite{zhang2020pegasus, nguyen2021enriching}. In addition to Monolingual Summarization, Neural Cross-Lingual Summarization has been receiving a tremendous amount of interest, likely due to the burgeoning need in cross-lingual information processing.

Conventional approaches designate a pipeline in two manners. The first one is translate-then-summarize, which copes with the task by initially translating the document into the target language and then performing the summarization \cite{wan2010cross, ouyang2019robust, wan2011using, zhang2016abstractive}. The second approach is summarize-then-translate, which firstly summarizes the document and then creates its translated version in the target language \cite{lim2004multi, orasan2008evaluation, wan2010cross}. Nonetheless, both of these approaches are vulnerable to error propagation caused by undertaking multiple steps \cite{zhu2019ncls}.

Recent works apply a general architecture combined with large-scale training to conduct Cross-Lingual Summarization. The main approach is to utilize the multi-task framework, in which CLS task benefits from the process of making use of other tasks such as Monolingual Summarization or Machine Translation \cite{zhu2019ncls}. Further approaches design ancillary mechanisms such as pointer-generator to exploit the translation scheme in the cross-lingual summary \cite{zhu2020attend}. Other work uses a pair of encoders and decoders to co-operate the cross-lingual alignment with summarization \cite{cao2020jointly}.

\subsection{Optimal Transport in Natural Language Processing}

Introduced in 19th century as a method to find the optimal solution to transport a mass from one place to another destination, researchers have found its use in a wide variety of scientific fields, such as computational fluid mechanics \cite{benamou2000computational}, economics \cite{carlier2015numerical}, physics \cite{cole2021quantum}, and notably machine learning \cite{peyre2019computational, cuturi2013sinkhorn, courty2016optimal, danila2006optimal}. 

Recently, beside Contrastive Learning framework \cite{nguyen2021contrastive, pan2021improved, pan2021contrastive}, Optimal Transport has been omnivorously employed in Natural Language Processing field, as used through Optimal Transport distance, for instance Word Mover's Distance \cite{werner2019speeding}, to estimate the necessary quantity of alignment. Its application includes text classification \cite{kusner2015word}, capturing spatial alignment in word embedding \cite{alvarez2018gromov}, machine translation \cite{chen2019improving}, abstractive summarization \cite{chen2019improving}, etc. Nevertheless, the adaptation of Optimal Transport distance, especially Sinkhorn divergence, for Neural Cross-Lingual Summarization task has been attracting limited amount of research effort.
\section{Background}

\subsection{Neural Cross-Lingual Summarization}

Given a document $X^{L_1} = \{x_1, x_2, …, x_N\}$, a monolingual summarization model’s task is to create a summary $Y^{L_1} = \{y^{L_1}_1, y^{L_1}_2, …, y^{L_1}_{M_1}\}$, where both $X^{L_1}$ and $Y^{L_1}$ are in language $L_1$. On the contrary, a cross-lingual summarization model will produce a cross-lingual summary $Y^{L_2} = \{y^{L_2}_1, y^{L_2}_2, …, y^{L_2}_{M_2}\}$ that is in language $L_2$. It is worth noting here that $M_1 < N$ and $M_2 < N$.

Analogous to monolingual summarization, current state-of-the-art cross-lingual summarization methods employ the Transformer-based architecture. Relying mainly on self-attention mechanism, Transformer-based architecture consists of an encoder and a decoder. The bidirectional self-attention in the encoder will extract contextualized representations of the input, which will be fed to the decoder to generate the output. Due to its generation nature, the decoder will use unidirectional self-attention to learn the context of previously generated tokens. During training procedure, the whole framework is updated based upon the cross-entropy loss as follows

\begin{equation}
    \mathcal{L}_{\text{CLS}} = -\sum_{t=1}^{M_2} \log P (y^{L_2}_{t} | y^{L_2}_{<t}, X^{L_1})
\end{equation}

\subsection{Knowledge Distillation (KD)}

Proposed by \cite{hinton2015distilling}, knowledge distillation is a method to train a model, called the student, by leveraging valuable information provided by soft targets output by another model, called the teacher. In particular, the framework initially trains a model on one designated task to extract useful features. Subsequently, given a dataset $D = \{ (X_1, Y_1), (X_2, Y_2), … (X_{|D|}, Y_{|D|})\}$, where $|D|$ is the size of the dataset, the teacher model will generate the output $H^T_i = \{\mathbf{h}^T_1,  \mathbf{h}^T_2, …, \mathbf{h}^T_{L_T}\}$ for each input $X_i$. Dependent on the researchers’ decision, the output might be hidden representations or final logits. As a consequence, in order to train the student model, the framework will use a KD loss that discriminates the output of the student model $H^S_i = \{\mathbf{h}^S_1,  \mathbf{h}^S_2, …, \mathbf{h}^S_{L_S}\}$ given input $X_i$ from the teacher output $H^T_i$. Eventually, the KD loss for input $X_i$ will possess the form as follows

\begin{equation}
    \mathcal{L}_{\text{KD}} = dist (H^T_i, H^S_i)
\end{equation}

\noindent where $dist$ is a distance function to estimate the discrepancy of teacher and student outputs.

The explicated Knowledge Distillation framework has shown its efficiency in a tremendous number of tasks, such as Neural Machine Translation \cite{tan2019multilingual, wang2021selective, li2021data, sun2020knowledge}, Question Answering \cite{hu2018attention, arora2019knowledge, yang2020model}, Image Classification \cite{yang2020knowledge, chen2018knowledge, fu2020interactive}, etc. Nonetheless, its application for Neural Cross-Lingual Summarization has received little interest.
\section{Methodology}

\begin{figure*}[t]
\centering
\includegraphics[width=0.9\linewidth]{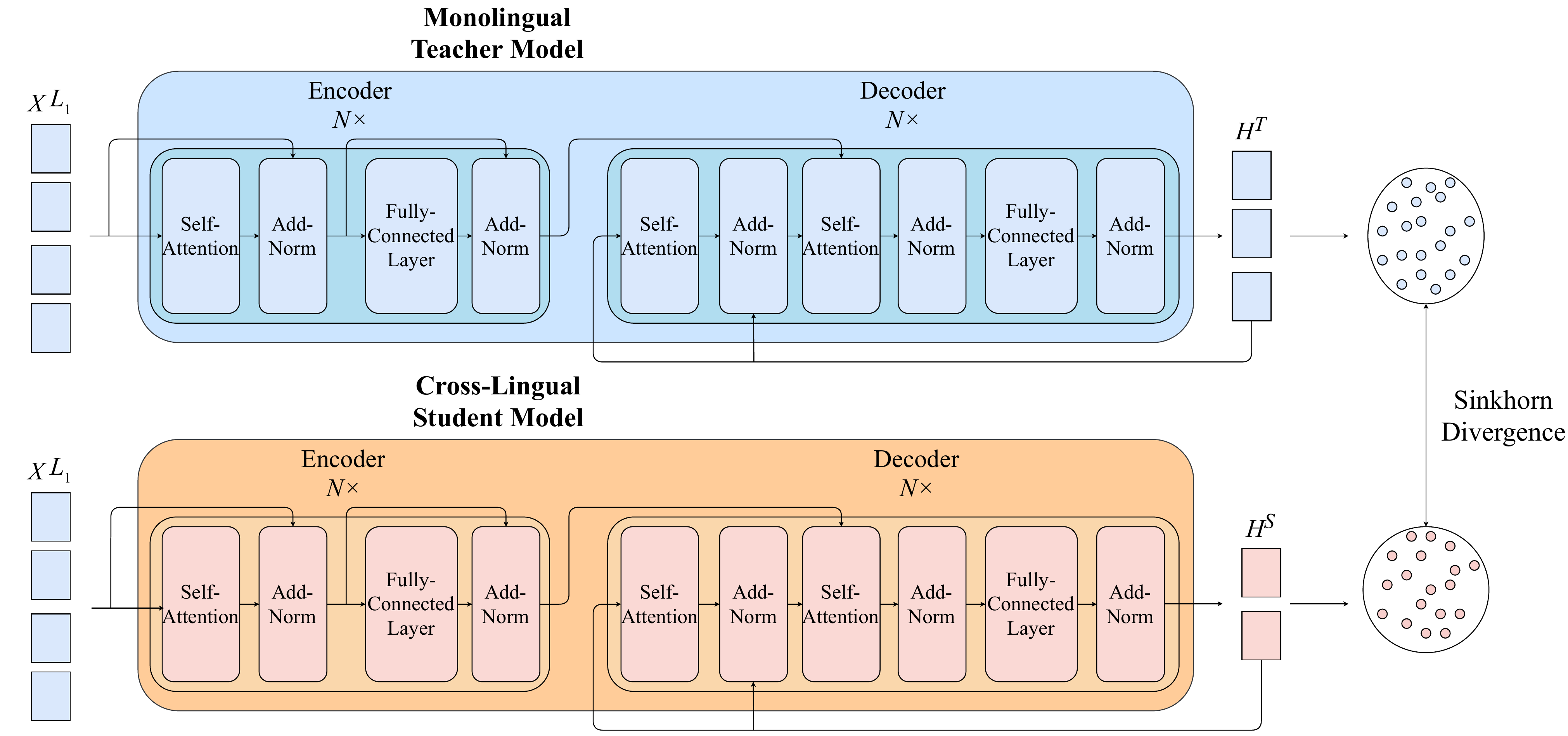}
\caption{Diagram of Knowledge Distillation Framework for Cross-Lingual Summarization}
\label{fig:model}
\end{figure*}

To resolve the issue of distant languages, the output representations from two vector spaces denoting two languages should be indistinguishable, or easily transported from one space to another. In order to accomplish that goal, we seek to relate the cross-lingual output of the student model to the monolingual output of the teacher model, via utilizing Knowledge Distillation framework and Sinkhorn Divergence calculation. The complete framework is illustrated in Figure \ref{fig:model}.

\subsection{Knowledge Distillation Framework for Cross-Lingual Summarization}
We inherit the architecture of Transformer model for our module. In particular, both the teacher and student model uses the encoder-architecture paradigm combined with two fundamental mechanisms. Firstly, the self-attention mechanism will attempt to learn the context of the tokens by attending tokens among each other in the input and output document. Secondly, there is a cross-attention mechanism to correlate the contextualized representations of the output tokens to ones of the input tokens. 

In our KD framework, we initiate the process by training the teacher model on monolingual summarization task. In detail, given an input $X^{L_1} = \{x_1, x_2, …, x_N\}$, the teacher model will aim to generate its monolingual summary $Y^{L_1} = \{y^{L_1}_1, y^{L_1}_2, …, y^{L_1}_{M_1}\}$. Similar to previous monolingual summarization schemes, our model is trained by maximizing the likelihood of the groundtruth tokens, which takes the cross-entropy form as follows

\begin{equation}
    \mathcal{L}_{\text{MLS}} =  -\sum_{t=1}^{M_1} \log P (y^{L_1}_t | y^{L_1}_{<t}, X^{L_1})
\end{equation}

After finetuning the teacher model, we progress to train the student model, which also employs the Transformer architecture. Contrary to the teacher, the student model’s task is to generate the cross-lingual output $Y^{L_2} = \{y^{L_2}_1, y^{L_2}_2, …, y^{L_2}_{M_2}\}$ in language ${L_2}$, given the input document $X^{L_1}$ in language ${L_1}$. We update the parameters of the student model by minimizing the objective function that is formulated as follows

\begin{equation}
    \mathcal{L}_{\text{CLS}} =  -\sum_{t=1}^{M_2} \log P (y^{L_2}_t | y^{L_2}_{<t}, X^{L_1})
\end{equation}

With a view to pulling the cross-lingual and monolingual representations nearer, we implement a KD loss to penalize the large distance of two vector spaces. Particularly, let $H^T = \{\mathbf{h}^T_1,  \mathbf{h}^T_2, …, \mathbf{h}^T_{L_T}\}$ denote the contextualized representations produced by the decoder of the teacher model, and $H^S = \{\mathbf{h}^S_1,  \mathbf{h}^S_2, …, \mathbf{h}^S_{L_S}\}$ denote the representations from the decoder of the student model, we define our KD loss as follows

\begin{equation}
    L_{\text{KD}} = dist(H^T, H^S)
\end{equation}

where $dist$ is the Optimal-Transport distance to evaluate the difference of two representations, which we will delineate in the following section.

\subsection{Sinkhorn Divergence for Knowledge Distillation Loss}
Due to the dilemma that the hidden representations of the teacher and student model stay upon two disparate vector spaces (as they represent two different languages), we will consider the distance of the two spaces as the distance of two probability measures. To elaborate, we choose to adapt Sinkhorn divergence, a variant of Optimal Transport distance, to calculate the aforementioned spatial discrepancy. Let $H^T$, $H^S$ denote the representations of the teacher decoder and the student decoder, we encode the sample measures of them

\begin{equation}
    \alpha =\sum_{i=1}^{L_T} \alpha_{i} \delta_{\mathbf{h}_i^T}, \quad \beta=\sum_{j=1}^{L_S} \beta_{j} \delta_{\mathbf{h}_j^S}
\end{equation}
where $\alpha$ and $\beta$ are probability distributions that satisfy $\sum_{i=1}^{L_T} \alpha_{i} = 1$ and $\sum_{j=1}^{L_S} \beta_{j} = 1$.

Inspired by \cite{feydy2019interpolating}, we estimate the difference of the representations through determining the Sinkhorn divergence between them

\begin{equation}
    dist(H^T, H^S) = \text{OT} (\alpha, \beta) - \frac{1}{2} \text{OT} (\alpha, \alpha)  - \frac{1}{2} \text{OT} (\beta, \beta) 
\end{equation}
where 
\begin{equation}
    \text{OT} (\alpha, \beta) = \sum_{i=1}^{N} \alpha_i f_i + \sum_{j=1}^{M} \beta_j g_j \\
\end{equation}
in which $f_i, g_j$ are estimated by Sinkhorn loop. We thorougly delineate the loop in Algorithm \ref{alg:sinkhorn_loop}. 

\begin{algorithm}[t]
\caption{Sinkhorn loop}
\label{alg:sinkhorn_loop}
\begin{algorithmic}[1]
\Require{Probability distributions $\alpha, \beta$, regularization hyperparameter $\varepsilon$, number of iterations $N_I$, log-sum-entropy function $\text{LSE}_{k=1}^{N} (z_k) = \log \sum_{k=1}^{N} \exp{(z_k)}$, distance function $C(\mathbf{x}, \mathbf{y}) = ||\mathbf{x} - \mathbf{y}||^{2}$}

\For{$i=1$ to $N_I$}
    \State Compute $f_i = \varepsilon \cdot \text{LSE}_{k=1}^{L_S} [\log (\beta_k) + \frac{1}{\varepsilon} g_k - \frac{1}{\varepsilon} C(\mathbf{h}^{T}_{i}, \mathbf{h}^{S}_{k})]$
    
    \State Compute $g_j = \varepsilon \cdot \text{LSE}_{k=1}^{L_T} [\log (\alpha_k) + \frac{1}{\varepsilon} f_k - \frac{1}{\varepsilon} C(\mathbf{h}^{T}_{k}, \mathbf{h}^{S}_{j})]$
\EndFor

\end{algorithmic}
\end{algorithm}

\subsection{Training Objective}
We amalgamate the Cross-Lingual Summarization and Knowledge Distillation objective to obtain the ultimate objective function. Mathematically, for each input, our training loss is computed as follows

\begin{equation}
    \mathcal{L} = \mathcal{L}_{\text{CLS}} + \lambda \cdot \mathcal{L}_{\text{KD}} 
    \label{eq:objective_function}
\end{equation}
where $\lambda$ is the hyperparameter that controls the influence of the cross-lingual alignment of two vector spaces.
\section{Experiments}

\subsection{Datasets}

We evaluate the effectiveness of our methods on En2Zh and Zh2En datasets processed by \cite{bai2021cross}. We also inherit their minimum, medium, and maximum settings in order to verify the effectiveness of our method under limited-resourced settings. The sample size under each setting is depicted in Table \ref{tab:low_resource_scenario_sample_size}. Furthermore, to further evaluate the performance of our method in various languages, we also preprocess datasets of Wikilingua \cite{ladhak2020wikilingua} in the manner that every sample is converted to a triple of document, MLS summary, and CLS summary. We choose 4 variants of Wikilingua to proceed our evaluation, i.e. English to Arabic (En2Ar), English to Japanese (En2Ja), Japanese to English (Ja2En), and English to Vietnamese (En2Vi). It should be noted here that (En, Ja), (En, Ar), (En, Zh), and (En, Vi) are all couples of languages that are distant in terms of structure or morphology. The statistics of the datasets is demonstrated in Table \ref{tab:dataset_statistics}. 

\begin{table}[H]
\centering
    \begin{tabular}{lrrr}
    \hline
    Dataset & $l_{\text{Input}}$ & $l_{\text{CLS}}$ & $l_{\text{MLS}}$ \\
    \hline
    Zh2En & 105 & 19 & 19 \\
    En2Zh & 912 & 97 & 69 \\
    En2Ar & 1589 & 227 & 133 \\
    En2Ja & 1463 & 212 & 133 \\
    Ja2En & 2103 & 133 & 212 \\
    En2Vi & 1657 & 175 & 135 \\
    \hline
    \end{tabular}
    \caption{Statistics of Cross-Lingual Summarization datasets.}
    \label{tab:dataset_statistics}
\end{table}

\begin{table}[H]
\centering
\small
    \begin{tabular}{lrrrr}
    \hline
    Scenarios & Minimum & Medium & Maximum & Full-dataset \\
    \hline
    Zh2En & 5,000 & 25,000 & 50,000 & 1,693,713 \\
    En2Zh & 1,500 & 7,500 & 15,000 & 364,687 \\
    \hline
    \end{tabular}
    \caption{Dataset sizes of multiple low-resource scenarios for CLS datasets.}
    \label{tab:low_resource_scenario_sample_size}
\end{table}

\subsection{Implementation Details}

We initialize the encoder with multilingual BERT \cite{devlin2018bert}, whereas the decoder with Xavier intialization \cite{glorot2010understanding}. The dimensions of  our encoder and decoder hidden states are both 768. We use two seperate Adam optimizers for encoder and decoder, and the learning rate for encoder and decoder is 0.002 and 0.2, respectively. The model is trained with the warmup phase of 25000 steps. We train the model on one Nvidia GeForce A100 GPU that accumulates gradient every 5 steps. Moreover, we apply Dropout probability of 0.1 to all fully-connected layers in the model. The teacher and student model shares the architecture and scale of parameters in our Knowledge Distillation framework. To estimate the Sinkhorn divergence, we employ the entropic regularization rate $\varepsilon$ of 0.0025 and the iteration length $N_I$ of 14. The weight $\lambda$ of KD Loss in Equation \ref{eq:objective_function} is set to 1.

\subsection{Baselines}
We compare our proposed architecture against the following baselines:
\begin{itemize}
    \item \textbf{NCLS} \cite{zhu2019ncls}: a Transformer-based model to conduct CLS.
    \item \textbf{NCLS + MS} \cite{zhu2019ncls}: a multi-task framework that leverages an auxiliary MS decoder to enhance cross-lingual summarization performance.
    \item \textbf{TLTran} \cite{bai2021cross}: a CLS pipeline that firstly performs MLS and then utilizes a finetuned NMT model to translate the monolingual summary into the target language.
    \item \textbf{MCLAS} \cite{bai2021cross}: a multi-task framework that sequentially performs MLS, and CLS which is based upon the MLS result.
\end{itemize}
\subsection{Automatic Evaluation}
\subsubsection{Full-dataset Scenario}
The experimental results under the full-dataset scenario are given in Table \ref{tab:full_cls_zh2en}, \ref{tab:full_cls_en2zh}, \ref{tab:full_cls_en2ar}, \ref{tab:full_cls_en2ja}, \ref{tab:full_cls_ja2en}, and \ref{tab:full_cls_en2vi}. 

For Zh2En dataset, our method outperforms MCLAS model by ROUGE-1 of 1.3 points, ROUGE-2 of 4.0 points, ROUGE-3 of 0.4 point, and ROUGE-L of 1.2 points. Our model also improves the performance of NCLS model for dataset En2Zh, with 0.6 point in ROUGE-1, 1.5 points in ROUGE-2, 0.1 point in ROUGE-3, and 0.8 point in ROUGE-L. For Arabic language, our model achieves the enhancement compared against NCLS model by 0.1 in ROUGE-1 score, 2.9 in ROUGE-2 score, 1.6 in ROUGE-3 score, and 5.1 in ROUGE-L score. In En2Ja dataset, we outperformed previous best method MCLAS by 0.6 point in ROUGE-1, 0.2 point in ROUGE-2, 0.2 point in ROUGE-3, and 0.5 point in ROUGE-L. 

Additionally, for the reverse dataset Ja2En, our method significantly achieves higher performance with the improvement of 1.0 point of ROUGE-1, 0.5 point of ROUGE-2, 0.4 point of ROUGE-3, and 0.4 point of ROUGE-L, compared with MCLAS model. Those results substantiate our hypothesis that our framework is able to enhance the capability of apprehending and summarizing a document into a summary of another distant language, since English alphabet does not have any character in common with Japanese, Arabic, and Chinese counterparts.

For En2Vi dataset, our method also obtains notable improvement over other state-of-the-art methods. As shown in Table \ref{tab:full_cls_en2vi}, our model outperforms MCLAS model by 0.1 in ROUGE-1, 2.9 in ROUGE-2, 1.6 in ROUGE-3, and 5.1 in ROUGE-L. This demonstrates that our method is also capable of buttressing the model capacity in situations where two languages are slightly morphologically or structurally similar, since Vietnamese and English do share a number of characters in their alphabets. 

{\renewcommand{\arraystretch}{1.05}
\begin{table}[H]
\centering
\small
\begin{tabular}{c|c|c|c|c}
\hline
\textbf{Model} & \textbf{R1} & \textbf{R2} & \textbf{R3} & \textbf{RL} \\ \hline  
TLTran & 33.64 & 15.58 & 12.02 & 29.74 \\
NCLS & 35.60 & 16.78 & 12.57 & 30.27  \\ 
NCLS+MS & 34.84 & 16.05 & 12.28 & 29.47 \\
MCLAS & 35.65 & 16.97 & 12.78 & 31.14 \\ \hline
Our Model & \textbf{36.93} & \textbf{20.99} & \textbf{13.20} & \textbf{32.33} \\
\hline
\end{tabular}
\caption{Full-dataset Cross-Lingual Summarization results in Zh2En dataset}
\label{tab:full_cls_zh2en}
\end{table}}

{\renewcommand{\arraystretch}{1.05}
\begin{table}[H]
\centering
\small
\begin{tabular}{c|c|c|c|c}
\hline
\textbf{Model} & \textbf{R1} & \textbf{R2} & \textbf{R3} & \textbf{RL} \\ \hline  
TLTran & 30.20 & 12.20 & 11.79 & 27.02 \\
NCLS & 44.16 & 24.28 & 17.13 & 30.23 \\ 
NCLS+MS & 42.68 & 23.51 & 15.62 & 29.24 \\
MCLAS & 42.27 & 24.60 & 16.07 & 30.09 \\ \hline
Our Model & \textbf{44.75} & \textbf{25.76} & \textbf{17.20} & \textbf{31.05} \\
\hline
\end{tabular}
\caption{Full-dataset Cross-Lingual Summarization results in En2Zh dataset}
\label{tab:full_cls_en2zh}
\end{table}}

{\renewcommand{\arraystretch}{1.05}
\begin{table}[H]
\centering
\small
\begin{tabular}{c|c|c|c|c}
\hline
\textbf{Model} & \textbf{R1} & \textbf{R2} & \textbf{R3} & \textbf{RL} \\ \hline  
NCLS & 36.80 & 17.36 & 10.79 & 27.25 \\ 
NCLS+MS & 35.53 & 17.01 & 10.33 & 26.36 \\
MCLAS & 36.28 & 17.27 & 10.81 & 27.56 \\ \hline
Our Model & \textbf{36.89} & \textbf{20.28} & \textbf{12.40} & \textbf{32.38} \\
\hline
\end{tabular}
\caption{Full-dataset Cross-Lingual Summarization results in En2Ar dataset}
\label{tab:full_cls_en2ar}
\end{table}}

{\renewcommand{\arraystretch}{1.05}
\begin{table}[H]
\centering
\small
\begin{tabular}{c|c|c|c|c}
\hline
\textbf{Model} & \textbf{R1} & \textbf{R2} & \textbf{R3} & \textbf{RL} \\ \hline  
NCLS & 29.55 & 15.99 & 10.25 & 23.03 \\ 
NCLS+MS & 29.42 & 15.83 & 10.12 & 23.00 \\
MCLAS & 29.60 & 16.08 & 10.14 & 33.20 \\ \hline
Our Model & \textbf{30.21} & \textbf{16.27} & \textbf{10.46} & \textbf{23.90} \\
\hline
\end{tabular}
\caption{Full-dataset Cross-Lingual Summarization results in En2Ja dataset}
\label{tab:full_cls_en2ja}
\end{table}}

{\renewcommand{\arraystretch}{1.05}
\begin{table}[H]
\centering
\small
\begin{tabular}{c|c|c|c|c}
\hline
\textbf{Model} & \textbf{R1} & \textbf{R2} & \textbf{R3} & \textbf{RL} \\ \hline  
NCLS & 32.78 & 12.66 & 6.33 & 26.43 \\ 
NCLS+MS & 32.50 & 12.02 & 6.15 & 26.41 \\
MCLAS & 33.20 & 12.57 & 6.33 & 27.27 \\ \hline
Our Model & \textbf{34.21} & \textbf{13.08} & \textbf{6.70} & \textbf{27.63} \\
\hline
\end{tabular}
\caption{Full-dataset Cross-Lingual Summarization results in Ja2En dataset}
\label{tab:full_cls_ja2en}
\end{table}}

{\renewcommand{\arraystretch}{1.05}
\begin{table}[H]
\centering
\small
\begin{tabular}{c|c|c|c|c}
\hline
\textbf{Model} & \textbf{R1} & \textbf{R2} & \textbf{R3} & \textbf{RL} \\ \hline  
NCLS & 36.75 & 16.37 & 8.04 & 28.69 \\ 
NCLS+MS & 36.28 & 16.14 & 8.03 & 28.61 \\
MCLAS & 36.31 & 15.91 & 7.75 & 28.62 \\ \hline
Our Model & \textbf{37.38} & \textbf{16.20} & \textbf{8.09} & \textbf{28.97} \\
\hline
\end{tabular}
\caption{Full-dataset Cross-Lingual Summarization results in En2Vi dataset}
\label{tab:full_cls_en2vi}
\end{table}}

\subsubsection{Low-resource Scenario}
We denote results of the experiments conducted under minimum, medium, and maximum scenarios in Table \ref{tab:min_cls_results}, \ref{tab:med_cls_results}, and \ref{tab:max_cls_results}.

For the minimum setting, our model achieves the improvement over previous methods. In particular, we outperformed MCLAS model by 1.3 points of ROUGE-1, 0.5 point of ROUGE-2, 0.2 point of ROUGE-3, and 0.3 point of ROUGE-L in Zh2En dataset. For En2Zh dataset, we obtain an increase of 3.6 points in ROUGE-1, 0.6 point in ROUGE-2, 0.3 point in ROUGE-3, and 1.4 points in ROUGE-L.

Under the medium setting, the performance of our method is also higher than MCLAS model with 0.1 point in ROUGE-1, 1.1 points in ROUGE-2, 0.5 point in ROUGE-3, and 3.0 points in ROUGE-L. The improvement is more critical for dataset En2Zh with an increase of 3.0 in ROUGE-1, 1.9 in ROUGE-2, 0.6 in ROUGE-3, and 0.5 in ROUGE-L.

Last but not least, in maximum scenario, for dataset Zh2En, our gains compared against MCLAS model are 0.4 point in ROUGE-1, 0.4 point in ROUGE-2, 0.5 point in ROUGE-3, and 0.7 point in ROUGE-L. In dataset En2Zh, our improvements are 2.9 points in ROUGE-1, 0.3 point in ROUGE-2, 0.4 point in ROUGE-3, and 0.7 point in ROUGE-L.

Those aforementioned results have shown that our method is also capable of elevating the Cross-Lingual Summarization performance when the available training dataset is scarce.

\begin{table}[H]
\centering
\small
\resizebox{1.05\linewidth}{!}{
\begin{tabular}{c|cc}
\hline
\textbf{Models} & \textbf{Zh2En} & \textbf{En2Zh} \\ \hline  
NCLS & 20.93/5.88/2.47/17.58 & 34.14/12.45/4.38/21.20 \\
NCLS+MS & 20.50/5.45/2.22/17.25 & 33.96/12.38/4.36/21.07 \\ 
MCLAS & 21.03/6.03/2.68/18.16 & 32.03/13.17/4.28/21.17 \\ \hline
Our Model & \textbf{22.37/6.50/2.91/18.47} & \textbf{35.59/13.77/4.57/22.56} \\ 
\hline
\end{tabular}}
\caption{Minimum Cross-Lingual Summarization Results}
\label{tab:min_cls_results}
\end{table}

\begin{table}[H]
\centering
\small
\resizebox{1.1\linewidth}{!}{
\begin{tabular}{c|cc}
\hline
\textbf{Models} & \textbf{Zh2En} & \textbf{En2Zh} \\ \hline  
NCLS & 26.42/8.90/4.49/22.05 & 35.98/15.88/8.97/23.79  \\
NCLS+MS & 26.86/9.06/4.58/22.47 & 38.95/18.09/9.73/25.39 \\ 
MCLAS & 27.84/10.41/4.91/24.12 & 37.28/18.10/9.48/25.26 \\ \hline
Our Model & \textbf{27.97/11.51/5.37/27.16} & \textbf{40.30/20.01/10.05/25.79} \\ 
\hline
\end{tabular}}
\caption{Medium Cross-Lingual Summarization Results}
\label{tab:med_cls_results}
\end{table}

\begin{table}[H]
\centering
\small
\resizebox{1.1\linewidth}{!}{
\begin{tabular}{c|cc}
\hline
\textbf{Models} & \textbf{En2Zh} & \textbf{Zh2En} \\ \hline  
NCLS & 29.05/10.88/6.56/24.32 & 40.18/19.86/10.33/26.52  \\
NCLS+MS & 28.63/10.63/6.24/24.00 & 39.86/19.87/10.23/26.64 \\ 
MCLAS & 30.73/12.26/6.98/26.51 & 38.35/19.75/10.64/26.41 \\ \hline
Our Model & \textbf{31.08/12.70/7.45/27.16} & \textbf{41.24/20.01/11.00/27.06} \\ 
\hline
\end{tabular}}
\caption{Maximum Cross-Lingual Summarization Results}
\label{tab:max_cls_results}
\end{table}

\subsection{Human Evaluation}
Because automatic metrics do not completely betray the quality of the methods, we conduct further human evaluation for more precise assessment. To fulfil our objective, we design two tests in order to elicit human judgements in two manners.

In the first experiment, we present summaries generated by NCLS, MCLAS, our model, and the gold summary, then asked seven professional English speakers to indicate the best and worst summaries in terms of informativeness, faithfulness, topic coherence, and fluency. We randomly sampled 50 summaries from En2Vi dataset and 50 others from Ja2En dataset. The score of a model will be estimated as the percentage of times it was denoted as the best minus the percentage of times it was denoted as the worst.

For the second experiment, we decide to adapt Question Answering (QA) paradigm to our framework. For each sample, we create two independent questions that underscore the key information from the input document. Participants would read and answer each question as best as they could. The score of a system will be equal to the proportion of questions that the participants answer correctly. 

Fleiss’ Karpa scores of our experiments are shown in Table \ref{tab:inter_agreement}. It is obvious that the scores prove a strong inter-agreement among the participants.

The experimental results in Table \ref{tab:human_eval} indicate that our model generates summaries that are conducive to human judgements, and have more likelihood to preserve important content in the original documents than summaries of other systems.

{\renewcommand{\arraystretch}{1.05}
\begin{table}[H]
\centering
\begin{small}
\begin{tabular}{c|c|c}
\hline
\textbf{Test} & \textbf{Fleiss' Kappa} & \textbf{Overall Agreement} \\ \hline
Preference & 0.57 & 64.95\% \\ 
QA & 0.64 & 82.15\% \\
\hline
\end{tabular}
\end{small}
\caption{Fleiss' Kappa and Overall Agreement percentage of each human evaluation test. Higher score indicates better agreement.}
\label{tab:inter_agreement}
\end{table}}

\begin{table}[H]
\centering
\small
\begin{tabular}{c|c|c}
\hline
\textbf{Models} & \textbf{Preference Score} & \textbf{QA score} \\ \hline
NCLS & -0.123 & 51.11 \\
MCLAS & 0.169 & 59.26 \\
Our Model & 0.498 & 71.85 \\ \hline
Gold Summary & 0.642 & 95.52 \\
\hline
\end{tabular}
\caption{Human evaluation}
\label{tab:human_eval}
\end{table}

\subsection{Analysis on Distance Methods}
We compare our implemented Sinkhorn Divergence with other distance methods. Particularly, we perform the mean or max-pooling of the teacher and student hidden representations. Subsequently, we evaluate the teacher and student discrepancy via Cosine Similarity (CS) or Mean Squared Error (MSE) of two pooled vectors. We show the numerical results in Table \ref{tab:analysis_distance_methods}. The results demonstrate the superiority of Sinkhorn Divergence over other approaches. We hypothesize that those approaches do not efficaciously capture the geometry nature of cross-lingual output representations.
\begin{table}[H]
\centering
\small
    \begin{tabular}{lrrrr}
    \hline
    Distance Methods & R-1 & R-2 & R-3 & R-L \\
    \hline
    Mean-CS & 44.20 & 24.54 & 16.96 & 30.27 \\
    Mean-MSE & 44.14 & 24.27 & 16.22 & 30.19 \\
    Max-CS & 44.29 & 25.65 & 17.07 & 30.82 \\
    Max-MSE & 44.23 & 24.61 & 16.44 & 30.21 \\
    \hline 
    Our Method & \textbf{44.75} & \textbf{25.76} & \textbf{17.20} & \textbf{31.05} \\
    \hline
    \end{tabular}
    \caption{Results when applying different distance methods in En2Zh dataset under full-dataset setting.}
    \label{tab:analysis_distance_methods}
\end{table}

\subsection{Case Study}

Figure \ref{fig:example} demonstrates case studies on the summarization results of NCLS, MCLAS, and our method. It is noticeable that summaries generated by NCLS and MCLAS missed important terms denoted by monolingual ones, such as ``\emph{consumers}'', ``\emph{cost}'', ``\emph{underground}'', ``\emph{parking lot}'', etc. Consequently, those summaries unfortunately fail to mention the main idea of the document, particularly the comparison of parking prices in sample 1, and the announcement of ``\emph{sinopec}'' in sample 2. In contrast, our outputs cover almost all of them, closely related the cross-lingual to the monolingual summary. This shows that attracting CLS and MLS representations towards each other through Sinkhorn divergence helps CLS model grasp key information, which is an advantage of our proposed method.

\subsection{Impact of Sinkhorn Divergence on Geometric Distance of Cross-Lingual Representations}

We propose to adapt Sinkhorn Divergence to align the cross-lingual decoder hidden states of the student model with monolingual decoder hidden states of the teacher model. Nevertheless, whether this geometrically brings two sets of representations nearer remains a quandary. To further verify the benefit of leveraging Sinkhorn Divergence, we estimate the distances of those hidden vectors by using other metrics, i.e. Cosine Similarity and Mean Squared Error. Particularly, for each input, after getting the decoder to generate the hidden vectors of the output tokens, we take the average of those vectors and measure the distance between the mean of the vectors generated by the CLS model (NCLS, MCLAS, and Our Model) with the mean of the vectors created by the MLS model. We denote the expected value and standard deviation of each method in Table \ref{tab:geometric_impact_of_sinkhorn_divergence}. As it can be obviously seen, employing Sinkhorn Divergence actually pulls the vectors in the cross-lingual spaces towards one another.

\begin{table}[H]
\centering
\small
    \begin{tabular}{lrr}
    \hline
    \textbf{Models} & \textbf{Cosine Similarity} & \textbf{Mean Squared Error} \\
    \hline
    NCLS & 0.165 $\pm$ 0.038 & 19.434 $\pm$ 7.252 \\
    MCLAS & 0.064 $\pm$ 0.057 & 17.207 $\pm$ 4.028 \\ \hline
    Our Model & \textbf{0.034} $\pm$ \textbf{0.054} & \textbf{13.517} $\pm$ \textbf{4.013} \\
    \hline
    \end{tabular}
    \caption{Results when applying different distance methods in Zh2En dataset under full-dataset setting.}
    \label{tab:geometric_impact_of_sinkhorn_divergence}
\end{table}
\section{Conclusion}

In this paper, we propose a novel Knowledge Distillation framework to tackle Neural Cross-Lingual Summarization for morphologically or structurally distant languages. Our framework trains a monolingual teacher model, and then finetunes the cross-lingual student model which is distilled knowledge from the aforementioned teacher. Since the hidden representations of the teacher and student model lie upon two different lingual spaces, we continually proposed to adapt Sinkhorn Divergence to efficiently estimate the cross-lingual discrepancy. Extensive experiments show that our method significantly outperforms other approaches under both low-resourced and full-dataset settings.

\bibliography{aaai22}
\end{document}